\pdfoutput=1
\documentclass[11pt]{article}

\usepackage[]{acl}

\usepackage{times}
\usepackage{verbatim}
\usepackage{latexsym}
\usepackage{graphicx}

\usepackage[T1]{fontenc}

\usepackage[utf8]{inputenc}
\usepackage{amsmath}
\usepackage{cleveref}

\usepackage{microtype}
\usepackage{booktabs}
\usepackage{multirow}

\usepackage{inconsolata}
\newcommand\Tstrut{\rule{0pt}{2.6ex}}       
\title{Learning from Natural Language Explanations for \\ Generalizable
Entity Matching}

\author{Somin Wadhwa$^\diamondsuit$* \quad Adit Krishnan$^\clubsuit$ \quad  \textbf{Runhui Wang}$^{\clubsuit}$ \quad \\
\textbf{Byron C. Wallace}$^\diamondsuit$ \quad \textbf{Chris Kong}$^\clubsuit$ 
  \\ 
$^\diamondsuit$Northeastern University \\
$^\clubsuit$ Amazon \\
\texttt{\small \{wadhwa.s,b.wallace\}@northeastern.edu} \\
\texttt{\small \{aditkris,runhuiw,luyankon\}@amazon.com}}

\begin{document}
\maketitle
\begingroup\def\thefootnote{*}\footnotetext{ Work perfomed during internship at Amazon.}\endgroup
\begin{abstract}

Entity matching is the task of linking records from different sources that refer to the same real-world entity. Past work has primarily treated entity linking as a standard supervised learning problem. However, supervised entity matching models often do not generalize well to new data, and collecting exhaustive labeled training data is often cost prohibitive. Further, recent efforts have adopted LLMs for this task in few/zero-shot settings, exploiting their general knowledge. But LLMs are prohibitively expensive for performing inference at scale for real-world entity matching tasks. 

As an efficient alternative, we re-cast entity matching as a conditional generation task as opposed to binary classification. This enables us to ``distill'' LLM reasoning into smaller entity matching models via natural language explanations. This approach achieves strong performance, especially on out-of-domain generalization tests ($\uparrow$10.85\% F-1) where standalone generative methods struggle. We perform ablations that highlight the importance of explanations, both for performance and model robustness. 
\end{abstract}

\section{Introduction}


\emph{Entity matching}, also known as \emph{record linkage} or \emph{data deduplication}, 
refers to matching 
records from different sources which refer to the same 
underlying entity, 
in the absence of unique identifiers. 
This is a practically important task across a diverse set of domains, e.g., database management, healthcare, customer relationship management, and financial services; in such applications, normalizing entities to realize a unified view of data is imperative.

Most prior work on entity matching has adopted supervised techniques, training a model to link entities within a particular domain. 
Performing pair-wise comparison on all record pairs is computationally prohibitive, especially on large scale datasets; typical entity resolution pipelines therefore perform \emph{blocking} followed by \emph{matching} \cite{ditto,wang2023sudowoodo}. 
The 
former step entails identifying candidate record pairs which may reference the same entity, 
while in the latter one attempts to infer whether this candidate is indeed a match.

\begin{figure}
    \centering
\includegraphics[scale=1.2]{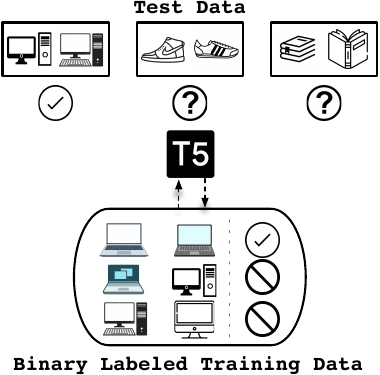}
\caption{An example of the generalization problem in entity matching: A model trained on a dataset of computers (e.g., WDC-Computers) is tested on instances taken from a corpus comprising shoes (WDC-Shoes).} 
\label{fig:intro}
\end{figure}

Assuming a supervised setting for this task 
is limiting in a few key ways.
First, collecting human supervision is inherently expensive. 
Second and relatedly, training an entity matching model in one ``domain'' (in this work, a domain is a product category) via explicit supervision will yield a model which is unlikely to readily transfer to other domains. 
For example, a model trained to match camera models based on descriptions is unlikely to generalize well to linking laptops (nevermind non-electronics). 
But collecting annotations linking products in all possible categories is not feasible. 
This has motivated work on transferable models for entity matching across domains  \cite{trabelsi2022dame,tu2022dader,tu2022domain,chai2023demystifying}.

One way to address the generalization problem may be to use general-purpose LLMs ``zero-shot'', via prompting and/or lightweight fine-tuning. 
Given the generality of such models, it is intuitive that they may be more robust to domain shifts when matching entities.
Moreover, an as-yet unexplored potential benefit of LLMs for this task is their ability to provide (natural language) ``reasoning'' for their outputs; this may permit fast manual verification of linkages, and therefore instill confidence in model outputs.
Aside from this, we later show that the richer signal in generated label ``rationales'' (or \textit{explanations}) allows for improved model distillation, consistent with recent findings on other tasks \cite{ho2022large}. 


A downside of LLMs is inference cost; applying such models to very large datasets---and continuously to new data as it is produced---is expensive. 
A comparatively tiny database with just one-thousand entities can yields a million (1k $\times$ 1k) candidate pairs, translating to thousands of dollars in inference costs.\footnote{\url{openai.com/pricing}}
We therefore explore \emph{model distillation for entity matching}.
In particular, we elicit ``reasoning'' alongside outputs for entity matching tasks from massive LLMs, and use this to train a modestly sized LM for entity matching such that it can also provide supporting rationales.\footnote{This is a type of distillation, but differs from traditional approaches \cite{Hinton2015DistillingTK} in that we are distilling only ``reasoning'' abilities, and not capabilities on the task itself.} 
We show that despite its small size, the resultant model achieves strong performance.
Moreover, our ablations highlight the importance of rationalization for robust entity matching, i.e., generalization. 

Our contributions are as follows.
(1) We frame entity matching as a conditional generation task and show that relatively small seq2seq models perform comparably to non-generative models when tested on in-domain instances. However, both approaches suffer significant loss in performance when tested on out-of-domain instances. 
(2) We show how augmenting entity matching training datasets with chain-of-thought style reasoning (explanations) obtained from larger models results in significant gains on out-of-domain instances. 
(3) We perform comprehensive ablations on LLM-generated ``explanations'' to 
    tease out which aspects of these explanations affect downstream model performance. 
    These findings may have implications for other tasks.  

\begin{table}[]
    \centering
    \scriptsize
    \begin{tabular}{@{}lccc@{}}
    \toprule
                   & \multicolumn{1}{c}{\begin{tabular}[c]{@{}c@{}}\textbf{Flan-T5}\\ \textit{(base)}\end{tabular}} & \multicolumn{1}{c}{\begin{tabular}[c]{@{}c@{}}\textbf{DITTO}\\ \textit{(RoBERTa-base)}\end{tabular}} & \multicolumn{1}{c}{\begin{tabular}[c]{@{}c@{}}\textbf{Mistral-7B LLM}\\ \textit{(Instruct)}\end{tabular}} \\     \midrule
    Training Method & \textbf{Supervised} & \textbf{Supervised} & \textbf{ICL Few-shot} \\
    \midrule
    Abt-Buy        &      \textbf{89.92}                                                                         &          89.33                &         31.11                                                       \\
    Amazon-Google  &       \textbf{76.23}                                                                        &            75.58               &        25.54                                                       \\
    Walmart-Amazon &     \textbf{87.40}                                                                          &              86.76              &       18.53                                                       \\
    Beer           &        93.33                                                                       &                \textbf{94.37}             &      32.91                                                       \\
    iTunes-Amazon  &      93.09                                                                         &                  \textbf{97.06}          &       41.88                                                       \\
    WDC-Computers  &    \textbf{92.08}                                                                           &            91.70                &       43.27                                                      \\
    WDC-Cameras    &       \textbf{91.25}                                                                        &          91.23                  &       45.31                                                       \\
    WDC-Watches    &        93.72                                                                       &          \textbf{95.69}                  &       53.94                                                       \\
    WDC-Shoes      &      \textbf{90.20}                                                                         &         88.07                  &        51.64                                                       \\ \bottomrule
    \end{tabular}
    \caption{Comparison of performance (F-1 scores) for prior work \cite{ditto} with recent generative models \cite{chung2022scaling} under full supervision (except on Mistral-7B LLM) on the task of entity matching under binary labeled (BL) data.}
    \label{tab:baselines}
    \end{table}

\begin{figure*}[h]
\centering
  \includegraphics[scale=0.70]{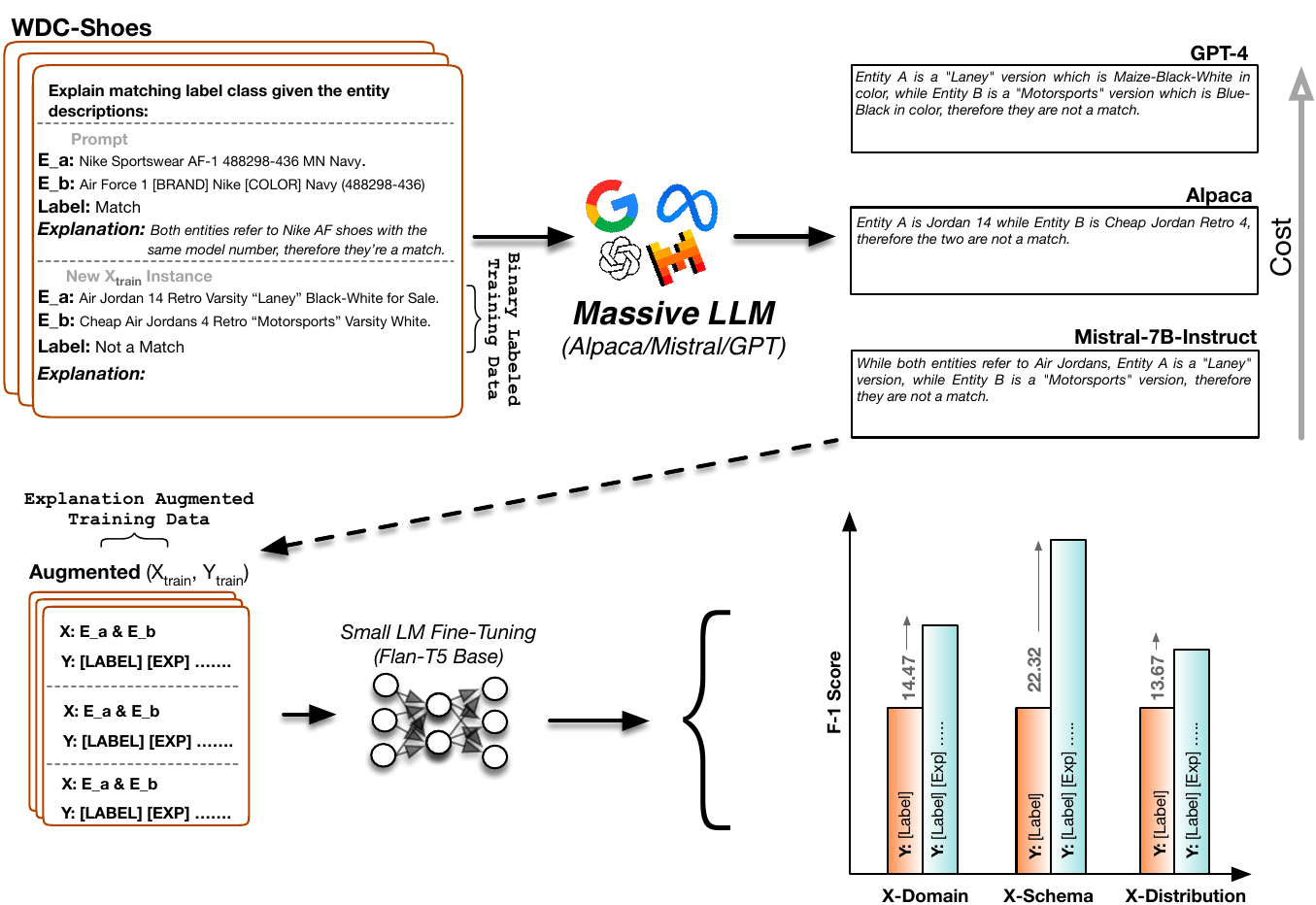}
  \caption{We propose augmenting binary labeled (BL) training data of entity matching datasets with Chain-of-Thought style natural language explanations from large models before fine-tuning smaller, more robust generative models. We use the time needed to generate explanation-augmented (EA) training data on a typical Amazon EC2 P3 instance as a proxy for cost in case of Mistral \cite{jiang2023mistral} and Alpaca \cite{alpaca} models, and the total cost of OpenAI's API usage in case of GPT-* models. Using this approach, we realize significant performance gains in a variety of out-of-domain test settings.}
  \label{fig:main}
\end{figure*}

\section{Entity Matching via Text Generation}
\label{sec:entity_matching_conditional_gen}
We treat entity matching as a conditional text generation task. For a dataset of $N$ entity pairs $x_i =$ \texttt{(entity$\_$a$_i$, entity$\_$b$_i$)}, we model the probability of generating classification label (e.g., "match"/"no match") as a string $y_i =$ $<y_{i}^{1}, y_{i}^{2} \cdots y_{i}^{T}>$, conditioned on a context string $\mathcal{C}_i$. Formally: 

\begin{equation*}
    p_{\text{LM}}(y_i | \mathcal{C}_i, x_i) = \prod_{t=1}^{T}p(y_i^{t} | \mathcal{C}_i, x_i, y_i^{1 \cdots t-1})
\end{equation*}

\noindent This is the standard conditional language modeling objective. During training, we use ``teacher-forcing'', i.e., 
condition production of outputs (``match'' or ``not'') on reference prefixes.

\subsection{Data}

We 
use $9$ publicly available entity matching datasets \cite{10.14778/1920841.1920904, 10.14778/2994509.2994535} used for evaluation in similar prior work \cite{ditto, peeters2023entity}. 
These datasets span several domains, allowing us to assess out-of-domain performance by testing a model trained on one type of data on examples from a another. 
Each dataset contains entity pairs from structured tables. 
We follow the \textit{input} linearization strategy and train/validation/test splits from \citet{ditto}.
Under this linearization scheme each input candidate entity pair is serialized as a sequence of tokens:

\begin{center}
    \texttt{[entity$_a$] [COL] $<$attr$>$ [VAL] ... \\}
    \texttt{[entity$_b$] [COL] $<$attr$>$ [VAL]...}
\end{center}

\noindent In our generative setting, a single training instance then becomes a pair of input text with entity attributes, and a linearized output target string\footnote{DITTO \cite{ditto} follows a non-generative approach and therefore does \textbf{not} require linearized strings as \textit{output targets}.}:

\begin{flushleft}
\small
\texttt{\textbf{Input }[entity$_a$] [COL] $<$Title$>$ [VAL] Nike Air Jordans 2007 ...  [entity$_b$] [COL] $<$Title$>$ Air Jordans by Nike [COL] $<$MANUF$\_$YEAR$>$ [VAL] 2007 ...}\\
\texttt{\textbf{Target } Match}
\end{flushleft}

\noindent We provide additional full length examples and dataset-specific instances in Appendix \ref{appx:datasets}.

\begin{table*}[ht!]
\centering
\small
\begin{tabular}{@{}lllrccc@{}}
\toprule
\multicolumn{1}{c}{\multirow{2}{*}{\textbf{Type}}} & \multicolumn{1}{c}{\multirow{2}{*}{\textbf{Training Data}}} & \multicolumn{1}{c}{\multirow{2}{*}{\textbf{Tested On}}} & \multicolumn{1}{c}{\multirow{2}{*}{\begin{tabular}[c]{@{}c@{}}\textbf{F-1}\\ (BL)\end{tabular}}} & \multicolumn{1}{c}{\multirow{2}{*}{\begin{tabular}[c]{@{}c@{}}\textbf{F-1}\\ (EA$_\text{Alpaca}$) \end{tabular}}} & \multicolumn{1}{c}{\multirow{2}{*}{\begin{tabular}[c]{@{}c@{}}\textbf{F-1}\\ (EA$_\text{Mistral}$) \end{tabular}}} &  \multicolumn{1}{c}{\multirow{2}{*}{$\nabla (\text{EA}_\text{Mistral-7B}-\text{BL})$ ($\uparrow$) }}                                                                              
\\
\multicolumn{1}{c}{}                               & \multicolumn{1}{c}{}                                        & \multicolumn{1}{c}{}                                    & \multicolumn{1}{c}{}                 & \multicolumn{1}{c}{}     & \multicolumn{1}{c}{}       & \multicolumn{1}{c}{}                                                                                            \\ \midrule
\multirow{15}{*}{\textbf{X-Domain}}                & Amazon-Google                                               & Beer                                                    & $70.27$                              &    $90.80$               & $92.30$                    & $22.03$                                                                                                                      \\
                                                   & Abt-Buy                                                     & Beer                                                    & $68.86$                              &    $85.11$               & $89.66$                    & $21.01$                                                                                                                   \\
                                                   & Walmart-Amazon                                              & Beer                                                    & $77.77$                              &    $85.62$               & $89.65$                    & $11.88$                                                                                                                     \\
                                                   & \multirow{3}{*}{WDC-Computers}                              & WDC-Shoes  \Tstrut                                      & $69.95$                              &    $76.16$               & $79.18$                    & $9.23$                                                                                                                    \\
                                                   &                                                             & WDC-Watches                                             & $80.07$                              &    $87.23$               & $87.02$                    & $6.94$                                                                                                                    \\
                                                   &                                                             & WDC-Cameras                                             & $73.26$                              &    $91.26$               & $93.77$                    & $20.57$                                                                                                                    \\
                                                   & \multirow{3}{*}{WDC-Shoes}                                  & WDC-Computers \Tstrut                                   & $67.90$                              &    $84.01$               & $84.13$                    & $16.23$                                                                                                                    \\
                                                   &                                                             & WDC-Watches                                             & $70.34$                              &    $81.49$               & $84.89$                    & $14.55$                                                                                                                     \\
                                                   &                                                             & WDC-Cameras                                             & $73.26$                              &    $82.27$               & $84.74$                    & $11.48$                                                                                                                   \\
                                                   & \multirow{3}{*}{WDC-Watches}                                & WDC-Computers \Tstrut                                   & $73.37$                              &    $85.43$               & $86.20$                    & $12.83$                                                                                                                     \\
                                                   &                                                             & WDC-Shoes                                               & $67.26$                              &    $80.99$               & $81.70$                    & $14.44$                                                                                                                    \\
                                                   &                                                             & WDC-Cameras                                             & $82.59$                              &    $88.47$               & $89.96$                    & $7.37$                                                                                                                \\
                                                   & \multirow{3}{*}{WDC-Cameras}                                & WDC-Computers \Tstrut                                   & $76.33$                              &    $86.92$               & $87.71$                    & $11.38$                                                                                                            \\
                                                   &                                                             & WDC-Watches                                             & $74.21$                              &    $80.20$               & $81.77$                    & $7.55$                                                                                                                    \\
                                                   &                                                             & WDC-Shoes                                               & $69.15$                              &    $78.52$               & $78.04$                    & $8.89$                                                                                                                    \\ \midrule
\multirow{4}{*}{\textbf{X-Schema}}                 & \multirow{2}{*}{iTunes-Amazon}                              & Amazon-Google                                           & $21.29$                              &    $43.45$               & $44.61$                    & $23.32$                                                                                                                     \\
                                                   &                                                             & Walmart-Amazon                                          & $20.04$                              &    $41.81$               & $43.09$                    & $23.05$                                                                                                                      \\
                                                   & Walmart-Amazon                                              & \multirow{2}{*}{iTunes-Amazon}                          & $51.72$                              &    $72.19$               & $75.63$                    & $23.91$                                                                                                                \\
                                                   & Amazon-Google                                               &                                                         & $72.22$                              &    $91.25$               & $91.21$                    & $18.99$                                                                                                                \\ \midrule
\multirow{9}{*}{\textbf{X-Distribution}}           & \multirow{2}{*}{Abt-Buy}                                    & Amazon-Google                                           & $22.25$                              &    $38.88$               & $41.42$                    & $19.17$                                                                                                                \\
                                                   &                                                             & Walmart-Amazon                                          & $25.77$                              &    $46.04$               & $45.09$                    & $19.32$                                                                                                      \\
                                                   & \multirow{2}{*}{Amazon-Google }                             & Abt-Buy  \Tstrut                                        & $26.72$                              &    $49.73$               & $44.64$                    & $17.92$                                                                                                          \\
                                                   &                                                             & Walmart-Amazon                                          & $33.10$                              &    $47.22$               & $51.61$                    & $18.51$                                                                                                              \\
                                                   & \multirow{2}{*}{Walmart-Amazon}                             & Abt-Buy   \Tstrut                                       & $63.75$                              &    $72.84$               & $67.52$                    & $3.77$                                                                                                               \\
                                                   &                                                             & Amazon-Google                                           & $52.05$                              &    $55.71$               & $60.20$                    & $7.97$                                                                                                                 \\
                                                   & \multirow{3}{*}{WDC-All}                                    & Abt-Buy \Tstrut                                         & $69.16$                              &    $76.58$               & $76.44$                    & $7.28$                                                                                                        \\
                                                   &                                                             & Amazon-Google                                           & $46.12$                              &    $56.12$               & $59.13$                    & $13.01$                                                                                                                 \\
                                                   &                                                             & Walmart-Amazon                                          & $64.09$                              &    $75.55$               & $76.37$                    & $12.28$                                                                                                              \\ \bottomrule
\end{tabular}
\caption{Comparison of FlanT5-base performance when trained without (BL) \textit{and} with explanation-augmented (EA) training data. Broadly, we observe significant gain in model performance when trained with chain-of-thought style explanations elicited from large language models.}
\label{tab:main_results}
\end{table*}

\subsection{Small LMs, SOTA Performance}
We start by evaluating baseline generative models to standard datasets. 
Table \ref{tab:baselines} summarizes our findings from these experiments. 
Generally, we find that even smaller generative models  (e.g., FlanT5-base) perform comparably to (and even occasionally outperform) their non-generative counterparts (e.g., DITTO). 
We also provide results from zero/ICL few-shot experiments using much larger generative models ($1$B$+$ parameters) in Appendix \ref{appx:llms}. 
 However, deploying such large models at scale would be prohibitively expensive. 
Therefore, we focus on smaller models in this work.

To quantify performance on \emph{out-of-domain} data, we 
consider three experimental settings representative of practical conditions under which entity matching models may be deployed. 
    
    \vspace{0.25em} \noindent \textbf{Cross Domain} 
    Train the model on entity pairs belonging to one domain (e.g., consumer electronics products) and test its performance on another domain (e.g., shoes). Training on the Amazon-Google dataset and testing model performance on WDC-Shoes is one example of this setting.

    \vspace{0.25em} \noindent \textbf{Cross Schema} 
    Entities in the test data may have different attributes, not seen in training, even if the data is from the same domain and derived from the same source. 
    Datasets used to test cross-schema robustness are \textit{not} mutually exclusive from (and may overlap with) cross-domain train-test data pairs. 
   
        \vspace{0.25em} \noindent \textbf{Cross Distribution} 
    Train and test the model on the same domain (e.g., consumer electronics products) but on entity pairs derived from different sources. For example: Train on Walmart-Amazon dataset, test  on the entity pairs of Abt-Buy data. 

\vspace{0.25em}
In every setting we observe, unsurprisingly, 
degraded model performance (F-1$_\text{(BL)}$ in Table \ref{tab:main_results}) compared to in-domain test sets (Table \ref{tab:baselines}). 
For instance, a model trained on a dataset of WDC-Cameras suffers a drop of $\sim$15 points when tested on a dataset of WDC-Computers. 
We provide additional results in Appendix \ref{appx:all_ood_results} for non-generative models under this cross testing framework. 
Broadly, consistent with prior work \cite{10.1145/3514221.3517870}, we find that non-generative models fare poorly when tested on out-of-domain data. 

We emphasize here that the aforementioned settings frequently occur and are a representative of the practical use-cases of entity matching models. It is often cost-prohibitive to collect and annotate data in large volumes for training domain, distribution, or schema-specific models. 

\subsection{Eliciting explanations from LLMs to improve smaller LMs}
\label{subsec:augment}
To improve out-of-domain model performance under our testing framework, we propose augmenting the binary labeled training data (BL) used to fine-tune small generative models with \textit{Chain-of-Thought} (CoT) style reasoning explanations \cite{Wei2022ChainOT} elicited from much larger language models Mistral-Instruct \cite{jiang2023mistral} and Alpaca \cite{alpaca}. We call this  explanation-augmented training data (EA). 

We use ICL few-shot prompting strategy to elicit meaningful generalizable CoT-style explanations given a pair of input entities and their corresponding matching label. 
Consider the following illustrative example from the WDC-Shoes dataset used as a prompt to elicit a \textit{CoT-explanation}. 

\begin{flushleft}
\small
\texttt{\textbf{Input }[entity$_a$] [COL] $<$Title$>$ [VAL] Nike Air Jordans 2007 ...  [entity$_b$] [COL] $<$Title$>$ Air Jordans by Nike [COL] $<$MANUF$\_$YEAR$>$ [VAL] 2007 ...}\\  
\texttt{\textbf{Target } Match \textcolor{red}{[explanation]} \textit{Both entities refer to Nike Air Jordans from 2007, therefore they're a match.}}
\end{flushleft}

\begin{flushleft}
\small
\texttt{\textbf{Input }[entity$_a$] [COL] $<$Title$>$ [VAL] New Balance 1080 Running [COL] $<$MANUF$\_$YEAR$>$ [VAL] 2016 ...  [entity$_b$] [COL] $<$Title$>$ NB Fresh Foam X 1080v13 [COL] $<$MANUF$\_$YEAR$>$ [VAL] 2016 ...}\\  
\texttt{\textbf{Target } Match \textcolor{red}{[explanation]} --}
\end{flushleft}

The actual prompts we use consist of two ICL examples (one for each target label type), in addition to the new instance for which we want the model to generate an explanation. 
An author of this paper wrote the explanations for the two ICL examples used in the prompt. We reproduce these prompts in their entirety in Appendix \ref{appx:prompts}. For generating \textit{CoT-style} explanations we used publicly available checkpoints for both Mistral-7B-Instruct\footnote{\url{huggingface.co/mistralai/Mistral-7B-Instruct-v0.1}} and Alpaca.\footnote{\url{crfm.stanford.edu/2023/03/13/alpaca.html}} We generated explanations with a maximum length of 128 tokens (minimum of 5 tokens) with top$_k$ sampling ($k=50$) and nucleus sampling ($p=0.95$). For every dataset, we found that generating explanations took approximately 2-5 seconds for Mistral-7B-Instruct, and 7-12 seconds on Alpaca-based models. 

We consider these model generated \textit{CoT-style} explanations analogous to summaries generated by a model given entity text and a corresponding matching label. 
We then use these explanations to fine-tune a smaller model (FlanT5-base in our case) and observe considerable gains in cross-domain, cross-schema, and cross-distribution performance (Table \ref{tab:main_results}). 
We find on average the F-1 score under cross-schema setting increases by $22.32$, while for cross-domain and cross-distribution setting the average F-1 score increases by $14.47$ and $13.67$ respectively.
In some instances (e.g., a model trained on \texttt{WDC-Computers} $\rightarrow$ tested on \texttt{WDC-Cameras}), we observe that augmenting the training set with \textit{CoT-style} explanations enables OOD performance comparable to in-domain performance\footnote{Details on reprehensibility are provided Appendix \ref{appx:reproducibility}.}.


\begin{table*}[t]
    \centering
    \small
    \begin{tabular}{@{}lllcrrrrr@{}}
    \toprule
    \multicolumn{1}{c}{\multirow{2}{*}{\textbf{Type}}} & \multicolumn{1}{c}{\multirow{2}{*}{\textbf{Training Data}}} & \multicolumn{1}{c}{\multirow{2}{*}{\textbf{Tested On}}} & \multicolumn{1}{c}{\multirow{2}{*}{\begin{tabular}[c]{@{}c@{}}\textbf{F-1}\\ (EA$_\text{Mistral}$)\end{tabular}}} & \multicolumn{5}{c}{\textbf{Ablations}}                                                                                \\
    \multicolumn{1}{c}{}                               & \multicolumn{1}{c}{}                                        & \multicolumn{1}{c}{}                                    & \multicolumn{1}{c}{}                                                                                             & \multicolumn{1}{c}{A} & \multicolumn{1}{c}{B} & \multicolumn{1}{c}{C} & \multicolumn{1}{c}{D} & \multicolumn{1}{c}{E} \\ \midrule
    \multirow{15}{*}{\textbf{X-Domain}}                & Amazon-Google                                               & Beer                                            & $92.30$                                                                                                              & $72.35$                 & $88.94$                 & $89.33$                 & $79.59$                 & $89.85$                 \\
                                                       & Abt-Buy                                                     & Beer                                            & $89.66$                                                                                                           & $62.99$                 & $88.81$                 & $87.93$                 & $70.01$                 & $87.50$                  \\
                                                       & Walmart-Amazon                                              & Beer                                            & $89.65$                                                                                                             & $75.25$                 & $89.30$                  & $91.47$                 & $76.29$                 & $83.33$                 \\
                                                       & \multirow{3}{*}{WDC-Computers}                              & WDC-Shoes  \Tstrut                                               & $79.18$                                                                                                            & $71.31$                 & $78.04$                 & $72.28$                 & $75.37$                 & $76.92$                 \\
                                                       &                                                             & WDC-Watches                                     & $87.01$                                                                                                            & $80.12$                 & $87.06$                 & $82.07$                 & $82.99$                 & $86.12$                 \\
                                                       &                                                             & WDC-Cameras                                     & $93.77$                                                                                                            & $69.15$                 & $91.92$                 & $89.86$                 & $88.56$                & $90.18$                \\
                                                       & \multirow{3}{*}{WDC-Shoes}                                  & WDC-Computers \Tstrut                                             & $84.13$                                                                                                            & $61.75$                 & $79.45$                 & $72.07$                 & $73.29$                 & $81.64$                    \\
                                                       &                                                             & WDC-Watches                                     & $84.89$                                                                                                            & $64.76$                 & $78.07$                 & $77.63$                 & $77.62$                 & $81.11$                    \\
                                                       &                                                             & WDC-Cameras                                     & $84.74$                                                                                                            & $72.23$                 & $77.61$                 & $74.95$                 & $77.03$                 & $82.61$                    \\
                                                       & \multirow{3}{*}{WDC-Watches}                                & WDC-Computers  \Tstrut                                            & $86.20$                                                                                                             & $78.18$                 & $84.64$                 & $84.99$                 & $76.05$                 & $85.71$                    \\
                                                       &                                                             & WDC-Shoes                                       & $81.70$                                                                                                            & $64.82$                 & $83.25$                 & $77.71$                 & $73.97$                 & $78.62$                    \\
                                                       &                                                             & WDC-Cameras                                     & $89.96$                                                                                                             & $85.92$                 & $89.36$                 & $88.61$                 & $85.25$                 & $89.18$                    \\
                                                       & \multirow{3}{*}{WDC-Cameras}                                & WDC-Computers   \Tstrut                                           & $87.71$                                                                                                           & $75.58$                 & $79.50$                  & $79.14$                 & $79.83$                 & $86.99$                 \\
                                                       &                                                             & WDC-Watches                                     & $81.77$                                                                                                            & $73.36$                 & $79.67$                 & $78.20$                  & $79.16$                & $77.21$                 \\
                                                       &                                                             & WDC-Shoes                                       & $78.04$                                                                                                            & $68.60$                  & $74.92$                 & $74.09$                 & $72.60$                  & $75.32$                                  \\ \midrule
    \multirow{4}{*}{\textbf{X-Schema}}                 & \multirow{2}{*}{iTunes-Amazon}                              & Amazon-Google                                             & $44.61$                                                                                                             & $20.89$                 & $32.44$                 & $35.57$                 & $35.58$                 & $35.05$                          \\
                                                       &                                                             & Walmart-Amazon                                  & $43.09$                                                                                                               & $17.14$                 & $40.49$                 & $39.08$                 & $41.16$                 & $25.64$                     \\
                                                       & Walmart-Amazon                                              & \multirow{2}{*}{iTunes-Amazon}                  & $75.63$                                                                                                              & $49.53$                 & $73.33$                 & $77.71$                 & $60.21$                 & $76.41$               \\
                                                       & Amazon-Google                                               &                                                 & $91.21$                                                                                                             & $69.56$                 & $83.65$                 & $83.23$                 & $73.07$                 & $89.97$   \\ \midrule
    \multirow{9}{*}{\textbf{X-Distribution}}           & \multirow{2}{*}{Abt-Buy}                                    & Amazon-Google                                           & $41.42$                                                                                                             & $24.73$                 & $36.56$                 & $42.04$                 & $27.76$                 & $39.64$                 \\
                                                       &                                                             & Walmart-Amazon                                  & $45.09$                                                                                                             & $22.01$                 & $44.09$                 & $43.84$                 & $27.84$                 & $40.75$                 \\
                                                       & \multirow{2}{*}{Amazon-Google}                              & Abt-Buy  \Tstrut                                                & $44.64$                                                                                                             & $23.31$                 & $32.05$                 & $45.08$                 & $31.29$                 & $33.61$                 \\
                                                       &                                                             & Walmart-Amazon                                  & $51.61$                                                                                                             & $29.55$                 & $35.47$                 & $42.54$                 & $36.55$                 & $45.08$                 \\
                                                       & \multirow{2}{*}{Walmart-Amazon}                             & Abt-Buy   \Tstrut                                                & $67.52$                                                                                                             & $62.81$                 & $68.99$                 & $68.11$                 & $64.91$                 & $67.55$                 \\
                                                       &                                                             & Amazon-Google                                   & $60.20$                                                                                                               & $51.92$                 & $60.47$                 & $58.83$                 & $54.27$                 & $58.84$                 \\
                                                       & \multirow{3}{*}{WDC-All }                                   & Abt-Buy \Tstrut                                                   & $76.44$                                                                                            & $68.48$                    & $71.28$                    & $72.36$                    & $70.21$                    & $75.51$                  \\
                                                       &                                                             & Amazon-Google                                   & $59.13$                                                                                                              & $49.74$                    & $55.49$                    & $55.12$                    & $50.56$                    & $53.99$                    \\
                                                       &                                                             & Walmart-Amazon                                  & $64.09$                                                                                                              & $62.19$                    & $73.81$                    & $72.43$                    & $67.23$                    & $75.28$                    \\ \midrule
    \multicolumn{3}{l}{$\nabla$ Aggregate comparison against F-1 (EA$_\text{Mistral}$)}                                                                                                           &                                                                                                                  & $-26.99$                    & $-5.57$                    & $-5.69$                    & $-14.35$                    & $-4.98$                    \\ \bottomrule
    \end{tabular}
    \caption{Comparison of FlanT5-base performance when LLM-generated explanations used during model training are ablated under various conditions -- \textbf{A.} Junk text substitution, \textbf{B.} Random reduction in length, \textbf{C.} TF-IDF reduction in length, \textbf{D.} Substitution with non-instance specific explanation, \textbf{E.} Random corruption of tokens in explanation.}
    \label{tab:ablations}
    \end{table*}

\section{Assessing the usefulness of explanations through ablations}
\label{section:ablations}

We conduct several ablations, both automated (labeled A--E) and through manual human annotations (H$_1$ and H$_2$), to assess the 
usefulness of generated explanations (which appear to improve the performance of \textit{smaller} entity-matching models). 
Table \ref{tab:ablations} summarizes findings from our automated ablations. 
We will use the following instance from the Abt-Buy dataset as a running example to demonstrate ablations A--E: 

\begin{flushleft}
\small
\texttt{\textbf{Entity A: }WD Red 3TB SATA III 3.5" Hard Drive - IntelliPower 64MB Cache WD30EFRX}\\
\texttt{\textbf{Entity B: }CCL Computers WD Red 1 - 64Mo (NAS) HDD} \\  
\texttt{\textbf{Label: }Not a Match} 
\end{flushleft}

For this instance, the language model (Mistral-7B-Instruct) generates the following explanation: 

\begin{flushleft}
\small
\texttt{\textbf{Generated: }While both entities refer to ``WD Red'' hard drive, Entity A specifically refers to 3TB SATA III 3.5" drive, while Entity B refers to a drive for use in a Network Attached Storage (NAS) and therefore they are not a match.}
\label{original}
\end{flushleft}

\noindent For each of the following ablations (A--E), we make targeted changes to the original LLM-generated explanations and then retrain the smaller LM 
to test the corresponding effects. 

\paragraph{A. Junk Substituion} We start by substituting LLM-generated explanations by sentences comprising random `junk' tokens, which are generated at random\footnote{via NLTK (\url{www.nltk.org})} from the English language vocabulary. 
We retain the original length of the explanation, e.g., in the example above the LLM-generated explanation is substituted with the following text

\begin{flushleft}
\small
\texttt{\textbf{Substituted:} contour fix nap egregious text nimble perhaps}
\end{flushleft}

\noindent The aim is to assess whether it is the presence of \emph{meaningful} text (rather than \textit{any} text) that leads to performance gains under the above settings. 
Aggregate performance under Ablation A drops $28.17\%$, and this is consistent across train-test pairs. 

\paragraph{B. Random Token-Drop} We alter the LLM-generated explanations by reducing their length. 
We start by removing all stop-words from the explanation, then randomly drop tokens to further reduce its length until we reduce the total length by half ($50\%$). 
In 
the running example, the LLM-generated explanation might be  replaced by the following text

\begin{flushleft}
\small
\texttt{\textbf{Substituted: }entities Red ``hard 3TB SATA 3.5'' use Attached Storage NAS match.}
\end{flushleft}

\paragraph{C. TF-IDF} 
Here we attempt to sample tokens from the LLM-generated explanation to assess if the presence of certain key tokens is all that is needed to realize the observed performance gains. 
We use TF-IDF \cite{salton1986introduction} as a measure of word importance. 
Specifically, we treat entity descriptions and their corresponding labels as \textit{documents}, and LLM-generated explanations as a \textit{summary} of these. 
We then sample tokens from the explanation based on the TF-IDF scores of individual tokens until we retain $50\%$ of the original length of the explanation. 
In the running example, the LLM-generated explanation might be replaced by the following text:
\begin{flushleft}
\small
\texttt{\textbf{Substituted: }drive to entity refers while 3tb and are attached both entities for hard iii in match nas network not red refer sata specifically storage}
\end{flushleft}

\noindent Perhaps surprisingly, 
sampling tokens in this way 
does \textit{not} help, compared to randomly sampling them like as in (B); the performance degradation is about the same ($5.57\%$ vs $5.69\%$; Table \ref{tab:ablations}). 

\paragraph{D. Generic Explanations} In this ablation we evaluate whether a \textit{dataset-level} (as opposed to instance-level) explanation yields performance gains. These dataset-wide explanations may or may not be model generated. 
For our experiments, we use the following manually written explanations:  
\begin{flushleft}
\small
\texttt{\textbf{WDC-Cameras }Based on the description of two cameras in Entity A and Entity B, they are \textit{(or are not)} a match.}\\
\texttt{\textbf{WDC-Shoes } Based on the color, brand, size and make of the two shoes in Entity A and Entity B respectively, they are \textit{(or are not)} a match.} \\ 
\texttt{\textbf{iTunes-Amazon } Based on the artist, genre and song titles, the two entities here are \textit{(or are not)} a match.}
\end{flushleft}

\noindent We find that the aggregate performance (Table \ref{tab:ablations}) declines by $\sim$14\%, compared to $\sim$25\% when we do not use any explanations, and $\sim$27\% using junk text as a substitute (Ablation A).  

\paragraph{E. Random Corruption} Finally, we evaluate the results when we randomly replace half of the tokens in LLM-generated explanation by a reserved token \texttt{($<$unk$>$)} to gauge whether the performance gains observed with explanations owe to the effective additional compute they permit at inference time. 
In our example, the LLM-generated explanation is modified to: 

\begin{flushleft}
\small
\texttt{\textbf{Substituted: }While \texttt{$<$unk$>$ $<$unk$>$ $<$unk$>$} to \texttt{$<$unk$>$ $<$unk$>$ $<$unk$>$ $<$unk$>$}' hard drive, \texttt{$<$unk$>$ $<$unk$>$} A specifically refers \texttt{$<$unk$>$}  3 \texttt{$<$unk$>$} SATA III \texttt{$<$unk$>$} 3.5 \texttt{$<$unk$>$ $<$unk$>$ $<$unk$>$ $<$unk$>$} ity B refers \texttt{$<$unk$>$ $<$unk$>$} drive \texttt{$<$unk$>$ $<$unk$>$ $<$unk$>$ $<$unk$>$ $<$unk$>$} Network \texttt{$<$unk$>$ $<$unk$>$} d \texttt{$<$unk$>$} (NAS) \texttt{$<$unk$>$} therefore \texttt{$<$unk$>$} are not \texttt{$<$unk$>$ $<$unk$>$} match \texttt{$<$unk$>$}}
\end{flushleft}

\noindent While we observe a performance difference on average (Table \ref{tab:ablations}), these differences are inconsistent across settings, contrary to our other ablation results. 
For instance, under cross-domain setting for \texttt{WDC-Cameras $\rightarrow$ WDC-Computers}, we observe that Ablation E outperforms both Ablations B and C and is comparable to using unaltered explanations. 
However, under a cross-schema setting for \texttt{iTunes-Amazon $\rightarrow$ Walmart-Amazon}, ablation E performs substantially worse than using unaltered explanations. We leave a more comprehensive analysis of this behavior for future work. 

In addition to ablations A--E, we conduct two additional experiments with human-interventions to test (1) robustness of models trained with augmented data; and (2) faithfulness of the generated reasoning explanations themselves. 
Because we generate tens of thousands of explanations (i.e., instance specific explanations for the entire training set for every dataset), collecting human annotations on all instances is cost prohibitive. 
Instead, we manually select $300$ instances from the Abt-Buy dataset to conduct the following two tests.

\begin{figure}
    \centering
\includegraphics[scale=0.63]{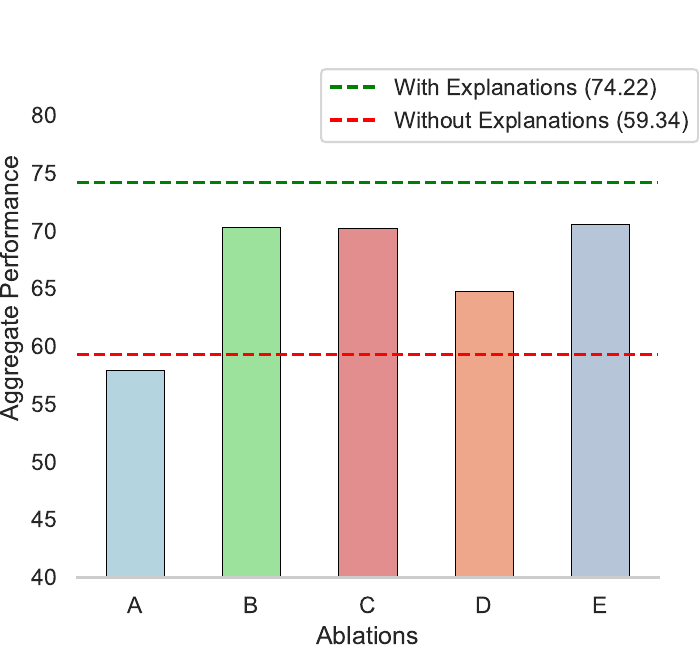}
\caption{Average F1 on out-of-domain test data when \textit{training} data is ablated under varying conditions.
} 
\label{fig:ablations_plot}
\end{figure}

\paragraph{H$_1$ Test of Robustness} First, we test robustness by randomly selecting 300 entity pairs with a ``match'' label from the test set. 
We then make minimal changes to the entity data (descriptions) to convert a ``matched'' to a ``non-matched'' pair. 
These changes are quite minimal, often involving only a token or two (e.g., \texttt{Nike}$\rightarrow$\texttt{Adidas}) while retaining a majority of token overlap between the entity pair descriptions. 
This intervention is motivated by the fact that matching models may over-rely on token overlap to classify whether or not the entity pair is match, and whether a trained model is robust to minor perturbations when tested on in-domain data. Consider the following example:

\begin{flushleft}
\small
\texttt{\textbf{Original: } [entity$_a$] Kingston 128GB  DataTraveler G3 USB 3.1  Flash drive [entity$_b$] Kingston 128G DT G3 USB 3.1 Flash Drive} \\
\texttt{\textbf{Label } Match} \\ 
\end{flushleft}

\begin{flushleft}
\small
\texttt{\textbf{Edited: }[entity$_a$] Kingston 128GB  DataTraveler G3 USB 3.1  Flash drive [entity$_b$] Kingston \textbf{32G} DT G3 USB 3.1 Flash Drive} \\
\texttt{\textbf{Corrected Label } Not a Match}
\end{flushleft}

\noindent Here we have minimally changed the storage capacity of two USB Flash Drives manufactured by the same company, under the same brand/model. 

We then run these substituted instances through our models -- trained both with \textit{\textbf{and}} without LLM-augmented explanations. Our goal here is was to test what percentage of labels correctly flip from ``match'' to ``no-match'' in both instances. We're motivated to test this aspect of robustness to determine the degree to which smaller trained models rely on raw token overlap vs the reasoning in LLM-generated explanations. 

For the models trained without explanations, we find that $71/300$ ($23\%$) of labels flip, while for the models trained with LLM-augmented explanations, we find that $164/300$ ($54\%$) labels successfully flip to a non-match; this indicates that augmented reasoning in training data makes smaller models more robust to subtle but critical input perturbations.

\paragraph{H$_2$ Test of Factuality} Finally, we investigate the extent to which LLM-generated explanations relate to the underlying entity pair descriptions. 
To this end we consider  generated explanations as analogous to document summaries, i.e., we consider the input entity pair descriptions and their matching label as a \textit{document}, and treat the model generated explanation of the \textit{summary}. 
We then annotate these explanations for inconsistencies. 

Three authors of this paper serve as human annotators and we use the Amazon Mechanical Turk (MTurk) sandbox as our preferred annotation platform. 
For every instance, we ask  annotators the following two questions related to the types of observed errors in reasoning explanations: 

\begin{flushleft}
\texttt{\textbf{Instrinsic Errors } Is the explanation fully derivable from the input entities and their corresponding matching label, irrespective of whether it contains excess information?} 
\end{flushleft}

\begin{flushleft}
\texttt{\textbf{Extrinsic Errors }  Does the explanation contain information in excess of the entity descriptions and their corresponding matching labels? These inconsistencies are often called ``hallucinations''.}
\end{flushleft}

We collected three annotations per instance and take the majority vote as reference where there is not unanimous agreement. 
We find that $10.9\%$ of instances contain instrinsic errors, and $15.1\%$ of explanations contain elements unsupported by inputs (``hallucinations''). 
We observe an inter-rater agreement (Fleiss's $\kappa$) of 0.75 for the question on instrinsic errors and an agreement of 0.86 on the question of extrinsic errors. 
We provide details on the annotation interface in Appendix \ref{appx:human_eval}. 

\section{Related Work}

\subsection{Deep learning in Entity Resolution}

With respect to entity resolution, the core process involves pairwise comparisons to ascertain matching entities. Recent efforts have capitalized on neural methods (including LLMs), including \textsf{DeepER} \cite{DBLP:journals/pvldb/EbraheemTJOT18}, a deep learning-based framework, and \textsf{DeepMatcher} \cite{DBLP:conf/sigmod/MudgalLRDPKDAR18}, which exemplifies the integration of deep learning in entity matching. 
Additionally, active learning strategies have been adapted for entity resolution as detailed in \cite{DBLP:conf/acl/KasaiQGLP19}. 

Other significant contributions include \texttt{Seq2SeqMatcher} \cite{nie2019deep, wang2024pre}, focusing on sequence-to-sequence matching, and \textsf{HierMatcher} \cite{fu2021hierarchical}, which adopts a hierarchical approach. The use of pre-trained language models has also gained traction, as evidenced by methods such as R-SupCon, Ditto, Rotom, and Sudowoodo, discussed in various studies \cite{brunner2020entity, DBLP:conf/vldb/PeetersBG20, ditto2021, DBLP:conf/sigmod/Miao0021, wang2022sudowoodo, wang2024neural, zeakis2023pre, genossar2023flexer}. These methods collectively represent the cutting-edge techniques in the realm of entity matching.

\textbf{Domain Adaptation} aims to allow a model trained in one domain to generalize to other domains \cite{trabelsi2022dame,tu2022dader,tu2022domain, sachidananda-etal-2021-efficient}. 

\subsection{Reasoning in LLMs}
Most recently, Entity Matching via LLMs has shown promising results \cite{peeters2023chatgpt, peeters2023llms, fan2024cost}. 
In these works, both zero-shot and fine-tuning approaches have been explored. Beyond entity matching, in-context learning (ICL) with LLMs has become a  dominant strategy, enabling these models to perform tasks with task conditioning and minimal task demonstrations \cite{ICL2brown2020language,ICL3xie2021explanation}. 
This approach has demonstrated strong performance \cite{ICL4zhao2021calibrate,ICL5liu2021makes} and streamlined experimentation with LLMs, as it eliminates the need for model training. 
However, the adoption of ICL has highlighted the sensitivity of LLMs to prompt selection \cite{ICL6lu2021fantastically,ICL7margatina2023active}, making prompt engineering for various tasks a challenging and time-consuming process. Nonetheless, data-driven signals, such as selecting semantically similar demonstrations using text retrievers, have proven to be effective \cite{ICL6lu2021fantastically,ICL7margatina2023active}, offering a more systematic approach to prompt engineering.

Chain-of-Thought (CoT) reasoning \cite{wang2022self,hoffmann2022training,chowdhery2022palm} has lately emerged as a means to allow LLMs to better perform certain tasks. 
This approach---which can be elicited via prompting few-shot examples \cite{kojima2022large}---involves guiding LLMs to generate a sequence of intermediate reasoning steps. 
Recent efforts have demonstrated the benefits of distilling ``reasoning'' capabilities in smaller LMs \cite{shridhar-etal-2023-distilling,wadhwa-etal-2023-revisiting}; our results contribute to this line of work. 

\section{Conclusions}
We proposed a novel model distillation approach to train a small, more-robust model for generalizable entity matching. 
Eliciting target label rationales from LLMs 
enables transfer of grounded ``reasoning'' to the smaller models. 
Our experiments show this translates to strong performance in diverse settings, outperforming existing models designed for domain adaptation that struggle to generalize. 
Ablation studies provide insight into the importance of explanation generation for achieving robust matching performance. 


\section*{Limitations}

We have shown that augmenting training data used to train smaller models with natural language explanations elicited from much larger models can yield substantial improvements in out-of-domain test settings. 
We then assessed the quality and usefulness of said explanations through automated ablations. 
Finally, we conducted human annotations on a sample of these explanations to quantify error they may contain. 

There are some important limitations to these findings. 
First, we have considered training a model on one domain (or distribution/schema), and then testing it on a set of \textit{$N-1$} datasets to evaluate model performance in an OOD setting. 
This (somewhat extreme) setting sharply exemplifies the sort of domain shift we are interested in studying. 
But we have not \textit{comprehensively} considered the more traditional OOD setting of training on \textit{$N-1$} datasets, and testing on the held out domain (distribution/schema), except while training on \texttt{WDC-All} and testing on \texttt{Abt-Buy, Amazon-Google}, and \texttt{Walmart-Amazon}. However, even under the limited circumstances we considered, we saw substantial gains in OOD performance ($\uparrow$10.86 F-1). 

Second, we rely on LLM-generated reasoning explanations to augment our training data. This dependence on externally hosted, proprietary large models could be problematic in certain sensitive domains, for example when working with entity descriptions that contain personally identifiable information (PII) since there is an extensive body of prior research \cite{hossain-etal-2023-misgendered, prakash-lee-2023-layered} documenting social biases inherent to LLMs.
That said, this dependence is only for \emph{training} data, and one could conceivably use open source LLMs, like we have, capable of CoT in place of proprietary models (e.g. OpenAI).

Third, while we find that distilling CoT-style explanations meaningfully improves small LM performance, our attempts to evaluating the usefulness of said explanations (if any) will require substantial future work. 
Our ablations do not provide a clear answer as to which aspects of these explanations are useful for downstream performance improvements. 
For instance, in ablation \textbf{D} we use a constant non-instance specific explanation appended to all target outputs (as opposed to instance specific explanation generated from a LLM). In theory, this provides no meaningful ability to classify a given instance over say, junk text. However, we still observe some gains in downstream OOD test performance.

Lastly, we \textit{only} experiment with datasets curated (and sourced) in English and therefore we do not have any insight into the issues that may result in other languages. 


\section*{Ethical Considerations}

\paragraph{Statement of Intended Use} Our work broadly relies on open-source datasets derived from e-commerce platforms, where entity attributes consist of heterogeneous descriptive sentences of common everyday consumer products. However, in certain applications of entity resolution like customer profile de-duplication, where entity descriptors involve human population-level attributes, the underlying data \textit{must} be appropriately de-identified (i.e. anonymized) in the interest of individual privacy. As stated in limitations, we make no attempt to manually edit/oversee the LLM-generated explanations before using them to train smaller LMs, and therefore there is a downstream risk of propagating large model biases. 
  


\bibliography{anthology,custom}
\clearpage
\appendix

\section*{Appendix}
\label{sec:appendix}

\section{Experimental settings and reproducibility}
\label{appx:reproducibility}
We performed all of our experiments on two AWS EC2 P3 instances, each containing 8 NVIDIA V100 (16GB) GPUs. We used the Huggingface library (v4.26.1; \citealt{wolf-etal-2020-transformers}) and publicly available checkpoints of models we used in our experiments. 
On all datasets except for WDC our best performing models were trained with batch size $16$, while for WDC datasets we used a batch size of $8$. 
We use default hyperparameters\footnote{\url{huggingface.co/docs/transformers/model_doc/flan-t5}} for model fine-tuning except for learning rate ($10^{-2} - 10^{-6}$), which we vary through hyperparameter tuning. We used the Adam optimizer and set the max epochs to $100$ with an early stopping patience of $10$ and a validation set F-1 score increase threshhold of $0.02$.
None of the trained models in any of our experiments required more than $60$ epochs. 

\section{Datasets}
\label{appx:datasets}
We select commonly used entity matching datasets in our work. Each dataset is split into training, validation, and test sets using the ratio 3:1:1 -- same splits as \citet{ditto} to provide direct comparisons in our OOD baselines (Table \ref{appxtab:baselines}):

\paragraph{Abt-Buy } This dataset contains product descriptions from e-commerce platforms \url{Abt.com} and \url{Buy.com}. A majority of products on either platform can be categorized as consumer electronics. There are a total of $9,575$ instances in the Abt-Buy dataset.

\paragraph{Amazon-Google } The Amazon-Google dataset consists mainly of software product offerings e.g. MS Office/Windows. The relevant entity attributes in Amazon-Google include \textit{brand, title} and \textit{price}. There are a total of $11,460$ product pairs. 

\paragraph{Walmart-Amazon } This is a structured benchmark entity matching dataset in the general consumer products domain containing textual product attributes like \textit{brand, title, model number,} and \textit{price}. Walmart-Amazon consists of $10,242$ product pairs.

\paragraph{iTunes-Amazon } Unlike our other datasets, iTunes-Amazon consists of strutured descriptions of songs in the form of textual attributes like \textit{artist, album year,} and \textit{title}. iTunes-Amazon is a relatively small dataset made up of $539$ instance pairs. 

\paragraph{Beer } This dataset contains structured textual attributes of beers from BeerAdvocate and RateBeer. We use the processed version\footnote{\url{pages.cs.wisc.edu/~anhai/data1/deepmatcher_data/Structured/Beer/exp_data}} of this dataset with the same train-dev-test splits as \citet{ditto}. There are only $450$ pairs in the Beer dataset. 

\paragraph{WDC Products } The Web Data Commons datasets span a variety of product categories like electronics, apparel, and accessories. WDC provides 4400 manually annotated gold labels from four categories: \texttt{computers ($68,461$), cameras ($42,277$), watches ($61,569$),} and \texttt{shoes ($42,989$)}. Each category contains 800 negative and 300 positive \textit{test} pairs. Each instance in all WDC datasets consists of four attributes - \textit{title, description, brand,} and \textit{specTable}. 

\section{Prompts}
\label{appx:prompts}
We use the following prompts as few-shot exemplars corresponding to each dataset \textit{type} to elicit natural language explanations. Inputs and target references are directly extracted from the original training sets while the \texttt{explanations} are human-written (by the authors) and were added for the experiments described in section \ref{subsec:augment}.

\paragraph{Consumer Electronic Products} We use the following prompt for all of the following datasets -- Abt-Buy, Amazon-Google, Walmart-Amazon, WDC-Computers, and WDC-Cameras. 
\begin{flushleft}
\small
\texttt{$<$s$>$$[$INST$]$ Given the following two examples, provide an explanation for the third example for why the two entities do or do not match. $[$\textbackslash INST$]$}  

\texttt{\textbf{Entity A:} [NAME] samsung dlp tv stand in black tr72bx [DESCRIPTION] samsung dlp tv stand in black tr72bx designed to fit samsung hlt7288 hlt7288 , hl72a650 , and hl67a650 television sets tempered 6mm tinted glass shelves wide audio storage shelves to accommodate 4 or more components wire management system easy to assemble high gloss black finish [PRICE] 369.0 }

\texttt{\textbf{Entity B:} [NAME] samsung tr72b tv stand [DESCRIPTION] glass black [PRICE] 232.14}  

\texttt{\textbf{Label:} MATCH}

\texttt{\textbf{Explanation:} Both entities refer to samsung TV stand in black and therefore have substantially similar specifications, therefore they're a match. $<$/s$>$}

\texttt{\textbf{Entity A:} [NAME] canon high capacity color ink cartridge color ink cl51 [DESCRIPTION] canon high capacity color ink cartridge cl51 compatible with pixma ip6210d , ip6220d , mp150 , mp170 and mp450 printers [PRICE] 35.0  }

\texttt{\textbf{Entity B:} [NAME] canon pg-40 twin pack black ink cartridge 0615b013 [DESCRIPTION] black [PRICE]}

\texttt{\textbf{Label:} NOT A MATCH}

\texttt{\textbf{Explanation:} Entity A refers to color ink cartridge while Entity B is a blank ink cartridge, therefore they are not a match. $<$/s$>$}
\end{flushleft}

\paragraph{Shoes} We use the following prompt for WDC-Shoes. The examples here are randomly selected from the WDC-Shoes training data. 
\begin{flushleft}
\small
\texttt{$<$s$>$ $[$INST$]$Given the following two examples, provide an explanation for the third example for why the two entities do or do not match.$[$/INST$]$}

\texttt{\textbf{Entity A:} [NAME]  Nike Sportswear Air Force 1 - Midnight Navy'en Mens Shoes Nike Navy 488298-436 en }

\texttt{\textbf{Entity B:} [NAME]  "Nike Air Force 1 '07 Low midnight navy / white (488298-436)"eu (488298-436) | Bludshop.com" eu}

\texttt{\textbf{Label:} MATCH}

\texttt{\textbf{Explanation:} Both entities refer to Nike Air Force shoes, navy in color with the same model number 488298-436, therefore they're a match.$<$/s$>$}

\texttt{\textbf{Entity A:} [NAME]  "Air Jordan 14 Retro Low “Laney” Varsity Royal/Varsity Maize-Black-White For Sale"en-US Sale | Cheap Jordans 2017"en-US }

\texttt{\textbf{Entity B:} [NAME]  "Cheap Air Jordan 4 Retro “Motorsports” White/Varsity Blue-Black Sale"en-US Sale | Cheap Jordans 2017"en-US }

\texttt{\textbf{Label:} NOT A MATCH}

\texttt{\textbf{Explanation:} While both entities refer to cheap Air Jordan shoes, Entity A is a Laney version which is Maize-Black-White in color, while Entity B is a Motorsports version which is Blue-Black in color, therefore they are not a match.$<$/s$>$}
\end{flushleft}

\paragraph{Music} We use the following prompt for iTunes-Amazon. The examples here are randomly selected from the iTunes-Amazon training data. 
\begin{flushleft}
\small
$<$s$>$ [INST] Given the following two examples, provide an explanation for the third example for why the two entities do or do not match. [\textbackslash INST]

\texttt{\textbf{Entity A:} [SONG\_NAME] Extra Extra Credit [ARTIST\_NAME] Wiz Khalifa [ALBUM\_NAME] Flight School [GENRE] Hip-Hop/Rap , Music [PRICE] 0.99 [COPYRIGHT] 2009 Rostrum Records [TIME] 4:03 [RELEASED] 17-Apr-09}

\texttt{\textbf{Entity B:} [SONG\_NAME] Extra Extra Credit [ Explicit ] [ARTIST\_NAME] Wiz Khalifa [ALBUM\_NAME] Flight School [ Explicit ] [GENRE] Rap \& Hip-Hop [PRICE] 0.99 [COPYRIGHT] 2013 Mad Decent [TIME] 4:03 [RELEASED] April 17 , 2009  }

\texttt{\textbf{Label:} MATCH}

\texttt{\textbf{Explanation:} Both entities are songs with the same name, artist and album.$<$/s$>$}

\textbf{Entity A:} [SONG\_NAME] Illusion ( feat . Echosmith ) [ARTIST\_NAME] Zedd [ALBUM\_NAME] True Colors [GENRE] Dance , Music, Electronic [PRICE] 1.29 [COPYRIGHT] 2015 Interscope Records [TIME] 6:30 [RELEASED] 18-May-15  

\texttt{\textbf{Entity B:} [SONG\_NAME] Papercut [ feat . Troye Sivan ] [ARTIST\_NAME] Zedd [ALBUM\_NAME] True Colors [GENRE] Dance \& Electronic [PRICE] 1.29 [COPYRIGHT] ( C ) 2015 Interscope Records [TIME] 7:23 [RELEASED] May 18 , 2015  }

\texttt{\textbf{Label:} NOT A MATCH}

\texttt{\textbf{Explanation:} While both entities refer to songs with the same artist, they have clearly different names and therefore, are not a match.$<$/s$>$}
\end{flushleft}

\paragraph{Beer} We use the following prompt for Beer dataset. 
\begin{flushleft}
\small
$<$s$>$ [INST] Given the following two examples, provide an explanation for the third example for why the two entities do or do not match.[\textbackslash INST]

\texttt{\textbf{Entity A:} [NAME] Honey Basil Amber [MANUFACTURER] Rude Hippo Brewing Company [STYLE] American Amber / Red Ale [ABV] 7.40 }

\texttt{\textbf{Entity B:} [NAME] Rude Hippo Honey Basil Amber [MANUFACTURER] 18th Street Brewery [STYLE] Amber Ale [ABV] 7.40 }

\texttt{\textbf{Label: }MATCH}

\texttt{\textbf{Explanation:} Both entities refer to Honey Basil Amber beer with the same ABV, therefore they're a match.$<$/s$>$}

\texttt{\textbf{Entity A:} [NAME] Brew Kahuna NW Red Ale [MANUFACTURER] Sky High Brewing [STYLE] American Amber / Red Ale [ABV] 5.20  } 

\texttt{\textbf{Entity B:} [NAME] Brew Bus Detour Series : Rollin Dirty Red Ale - Wood Aged [MANUFACTURER] Cigar City Brewing [STYLE] Irish Ale [ABV] 5 }

\texttt{\textbf{Label:} NOT A MATCH}

\texttt{\textbf{Explanation:} Entity A refers to Beer manufactured by Sky High Brewing while Entity B refers to Beer manufactured by Cigar City Brewing, and they have different names, therefore they are not a match.$<$/s$>$}
\end{flushleft}

\begin{table*}[ht!]
\centering
\small
\begin{tabular}{@{}lllcc@{}}
\toprule
\multicolumn{1}{c}{\multirow{2}{*}{\textbf{Type}}} & \multicolumn{1}{c}{\multirow{2}{*}{\textbf{Training Data}}} & \multicolumn{1}{c}{\multirow{2}{*}{\textbf{Tested On}}} & \multicolumn{1}{c}{\multirow{2}{*}{\begin{tabular}[c]{@{}c@{}}\textbf{F-1}\\ BL$_\text{DITTO}$ \end{tabular}}} &  \multicolumn{1}{c}{\multirow{2}{*}{\begin{tabular}[c]{@{}c@{}}\textbf{F-1}\\ BL$_\text{FlanT5-Base}$ \end{tabular}}}                                                                           
\\
\multicolumn{1}{c}{}                               & \multicolumn{1}{c}{}                                        & \multicolumn{1}{c}{}                                    & \multicolumn{1}{c}{}                 & \multicolumn{1}{c}{}                                                                                                \\ \midrule
\multirow{15}{*}{\textbf{X-Domain}}                & Amazon-Google                                               & Beer                                                    & $70.27$                              &    $63.10$                                                                                                                                   \\
                                                   & Abt-Buy                                                     & Beer                                                    & $68.86$                              &    $55.29$                                                                                                                                \\
                                                   & Walmart-Amazon                                              & Beer                                                    & $77.77$                              &    $59.12$                                                                                                                                 \\
                                                   & \multirow{3}{*}{WDC-Computers}                              & WDC-Shoes  \Tstrut                                      & $69.95$                              &    $65.18$                                                                                                                                  \\
                                                   &                                                             & WDC-Watches                                             & $80.07$                              &    $80.98$                                                                                                                               \\
                                                   &                                                             & WDC-Cameras                                             & $73.26$                              &    $70.51$                                                                                                                                 \\
                                                   & \multirow{3}{*}{WDC-Shoes}                                  & WDC-Computers \Tstrut                                   & $67.90$                              &    $65.11$                                                                                                                                \\
                                                   &                                                             & WDC-Watches                                             & $70.34$                              &    $74.47$                                                                                                                                \\
                                                   &                                                             & WDC-Cameras                                             & $73.26$                              &    $72.90$                                                                                                                             \\
                                                   & \multirow{3}{*}{WDC-Watches}                                & WDC-Computers \Tstrut                                   & $73.37$                              &    $75.34$                                                                                                                             \\
                                                   &                                                             & WDC-Shoes                                               & $67.26$                              &    $67.22$                                                                                                                               \\
                                                   &                                                             & WDC-Cameras                                             & $82.59$                              &    $81.16$                                                                                                                           \\
                                                   & \multirow{3}{*}{WDC-Cameras}                                & WDC-Computers \Tstrut                                   & $76.33$                              &    $75.83$                                                                                                                      \\
                                                   &                                                             & WDC-Watches                                             & $74.21$                              &    $73.92$                                                                                                                             \\
                                                   &                                                             & WDC-Shoes                                               & $69.15$                              &    $61.73$                                                                                                                             \\ \midrule
\multirow{4}{*}{\textbf{X-Schema}}                 & \multirow{2}{*}{iTunes-Amazon}                              & Amazon-Google                                           & $21.29$                              &    $21.48$                                                                                                                              \\
                                                   &                                                             & Walmart-Amazon                                          & $20.04$                              &    $18.75$                                                                                                                              \\
                                                   & Walmart-Amazon                                              & \multirow{2}{*}{iTunes-Amazon}                          & $51.72$                              &    $50.82$                                                                                                                       \\
                                                   & Amazon-Google                                               &                                                         & $72.22$                              &    $76.17$                                                                                                                        \\ \midrule
\multirow{9}{*}{\textbf{X-Distribution}}           & \multirow{2}{*}{Abt-Buy}                                    & Amazon-Google                                           & $22.25$                              &    $19.15$                                                                                                                      \\
                                                   &                                                             & Walmart-Amazon                                          & $25.77$                              &    $28.99$                                                                                                              \\
                                                   & \multirow{2}{*}{Amazon-Google }                             & Abt-Buy  \Tstrut                                        & $26.72$                              &    $25.55$                                                                                                               \\
                                                   &                                                             & Walmart-Amazon                                          & $33.10$                              &    $23.78$                                                                                                                             \\
                                                   & \multirow{2}{*}{Walmart-Amazon}                             & Abt-Buy   \Tstrut                                       & $63.75$                              &    $58.11$                                                                                                                            \\
                                                   &                                                             & Amazon-Google                                           & $52.05$                              &    $39.18$                                                                                                                              \\
                                                   & \multirow{3}{*}{WDC-All}                                    & Abt-Buy \Tstrut                                         & $69.16$                              &    $67.22$                                                                                                                    \\
                                                   &                                                             & Amazon-Google                                           & $46.12$                              &    $41.37$                                                                                                                           \\
                                                   &                                                             & Walmart-Amazon                                          & $64.09$                              &    $64.88$                                                                                                                     \\ \bottomrule
\end{tabular}
\caption{Comparison of OOD test performance under our framework for FlanT5-base \cite{chung2022scaling} and non-generative DITTO \cite{ditto} when trained on binary labeled (BL) training data. Broadly, we observe significant degradation in model performance under both models.}
\label{appxtab:baselines}
\end{table*}

\section{OOD Performance in Neural Entity Matching}
\label{appx:all_ood_results}
We conduct baseline experiments using our testing framework (cross-domain, cross-distribution, and cross-schema) on both generative (FlanT5) and non-generative (DITTO -- based on RoBERTa) methods. Table \ref{appxtab:baselines} summarizes our results. We observe significant decline in performance under both methods, with RoBERTa-based DITTO (Avg F-1: $55.28$) faring slightly worse than FlanT5 (Avg F-1: $59.28$). 

Our results on non-generative models like DITTO are in-line with prior work in the area where \citet{10.1145/3514221.3517870} first highlight the issue of domain adaptation and the challenge of \textit{reusing} labeled source data where there might be a change in distribution or domain at test time. 

\begin{table*}[]
\centering
\scriptsize
\begin{tabular}{@{}lrrrrrrrrrrrrrrr@{}}
\toprule
                        & \multicolumn{3}{c}{\textbf{Alpaca (7B)}}                                & \multicolumn{3}{c}{\textbf{Mistral-7B-Ins}}                        & \multicolumn{3}{c}{\textbf{Falcon-Ins (7B)}}                       & \multicolumn{3}{c}{\textbf{FlanT5-XXL }}                           & \multicolumn{3}{c}{\textbf{Flan-UL2}}                             \\ \cmidrule(l){2-16} 
\multicolumn{1}{c}{}    & \multicolumn{1}{c}{P} & \multicolumn{1}{c}{R} & \multicolumn{1}{c}{F-1} & \multicolumn{1}{c}{P} & \multicolumn{1}{c}{R} & \multicolumn{1}{c}{F-1} & \multicolumn{1}{c}{P} & \multicolumn{1}{c}{R} & \multicolumn{1}{c}{F-1} & \multicolumn{1}{c}{P} & \multicolumn{1}{c}{R} & \multicolumn{1}{c}{F-1} & \multicolumn{1}{c}{P} & \multicolumn{1}{c}{R} & \multicolumn{1}{c}{F-1} \\ \cmidrule(l){2-16} 
\textbf{A-B}        & 12.33                 & 77.61                 & 21.28                   & 16.49                 & 52.6                  & 25.11                   & 14.77                 & 50.81                 & 22.89                   & 15.23                 & 91.30                 & 26.11                   & 85.74                 & 42.41                 & 56.75                   \\
\textbf{A-G}  & 11.91                 & 89.29                 & 21.02                   & 15.50                 & 72.64                 & 25.54                   & 12.67                 & 70.41                 & 21.48                   & 20.75                 & 80.27                 & 32.98                   & 74.66                 & 48.3                  & 58.65                   \\
\textbf{W-A} & 10.31                 & 83.81                 & 18.37                   & 10.74                 & 75.40                 & 18.53                   & 11.52                 & 85.36                 & 20.30                   & 18.14                 & 72.09                 & 28.99                   & 92.21                 & 36.88                 & 52.69                   \\
\textbf{Beer}           & 18.91                 & 100.00                & 31.81                   & 20.01                 & 92.85                 & 32.91                   & 10.58                 & 100.00                & 19.14                   & 9.65                  & 89.30                 & 17.42                   & 13.5                  & 94.12                 & 23.61                   \\
\textbf{iT-A}  & 15.61                 & 95.66                 & 26.84                   & 28.32                 & 87.59                 & 42.80                   & 11.57                 & 98.47                 & 20.71                   & 15.46                 & 77.77                 & 25.79                   & 20.69                 & 85.12                 & 33.29                   \\
\textbf{W-Com}  & 29.74                 & 84.24                 & 43.96                   & 32.49                 & 64.76                 & 43.27                   & 29.59                 & 91.20                 & 44.68                   & 23.71                 & 82.45                 & 36.83                   & 92.55                 & 60.41                 & 73.10                   \\
\textbf{W-Cam}    & 30.57                 & 85.40                 & 45.02                   & 33.08                 & 72.24                 & 45.31                   & 26.99                 & 90.16                 & 41.54                   & 36.05                 & 87.77                 & 51.11                   & 80.51                 & 61.97                 & 70.03                   \\
\textbf{W-Wat}    & 35.49                 & 85.36                 & 50.14                   & 34.47                 & 75.68                 & 47.37                   & 11.17                 & 83.18                 & 19.70                   & 34.19                 & 85.44                 & 48.84                   & 84.13                 & 68.82                 & 75.71                   \\
\textbf{W-Sh}      & 32.79                 & 62.24                 & 42.95                   & 32.51                 & 78.35                 & 51.64                   & 36.43                 & 75.19                 & 49.08                   & 29.22                 & 65.09                 & 29.22                   & 75.48                 & 50.17                 & 60.28                   \\ \bottomrule
\end{tabular}
\caption{ICL Few Shot performance without any model training.}
\label{appxtab:llms}
\end{table*}

\section{Zero-Shot Entity Matching with LLMs}
\label{appx:llms}
In addition to training and testing smaller seq2seq models we also provide results from few-shot prompting on larger language models (\# parameters $>$ 7B). We emphasize here again that in \textit{any} practical entity matching context, deployment of such larger models is infeasible due the sheer number of comparisons involved. For instance, a \textit{small} product catalog of $1,000$ products can, in worst case scenario, lead to $1,000,000$ pair comparisons -- this requires efficiency and, as a practical matter, low deployment costs. Nevertheless, we feel it is important to contextualize our work under ICL few-shot settings on LLMs given their current relevance. We use the same prompts as provided in Appendix \ref{appx:prompts}, with one example of each class and test five \cite{alpaca, jiang2023mistral, falcon40b, chung2022scaling, tay2023ul2} instruction tuned models. 

Table \ref{appxtab:llms} summarizes these results. Generally, we find that all the models we test under-perform trained smaller LMs. We also observe certain behaviors while prompting LLMs where in some cases (see Alpaca tested on the Beer dataset) we get unusually high recall while getting very low precision measurements, indicating that models may excessively rely on token overlap as a proxy for entity matches. This is in line with prior work where \citet{peeters2023using} use ChatGPT for Entity Matching and observe similar behavior. We do not experiment with different prompts and/or chain-of-thought style explanations under these few-shot settings since that is beyond the scope of this work. 

\begin{figure*}[h]
\centering
  \includegraphics[scale=0.5]{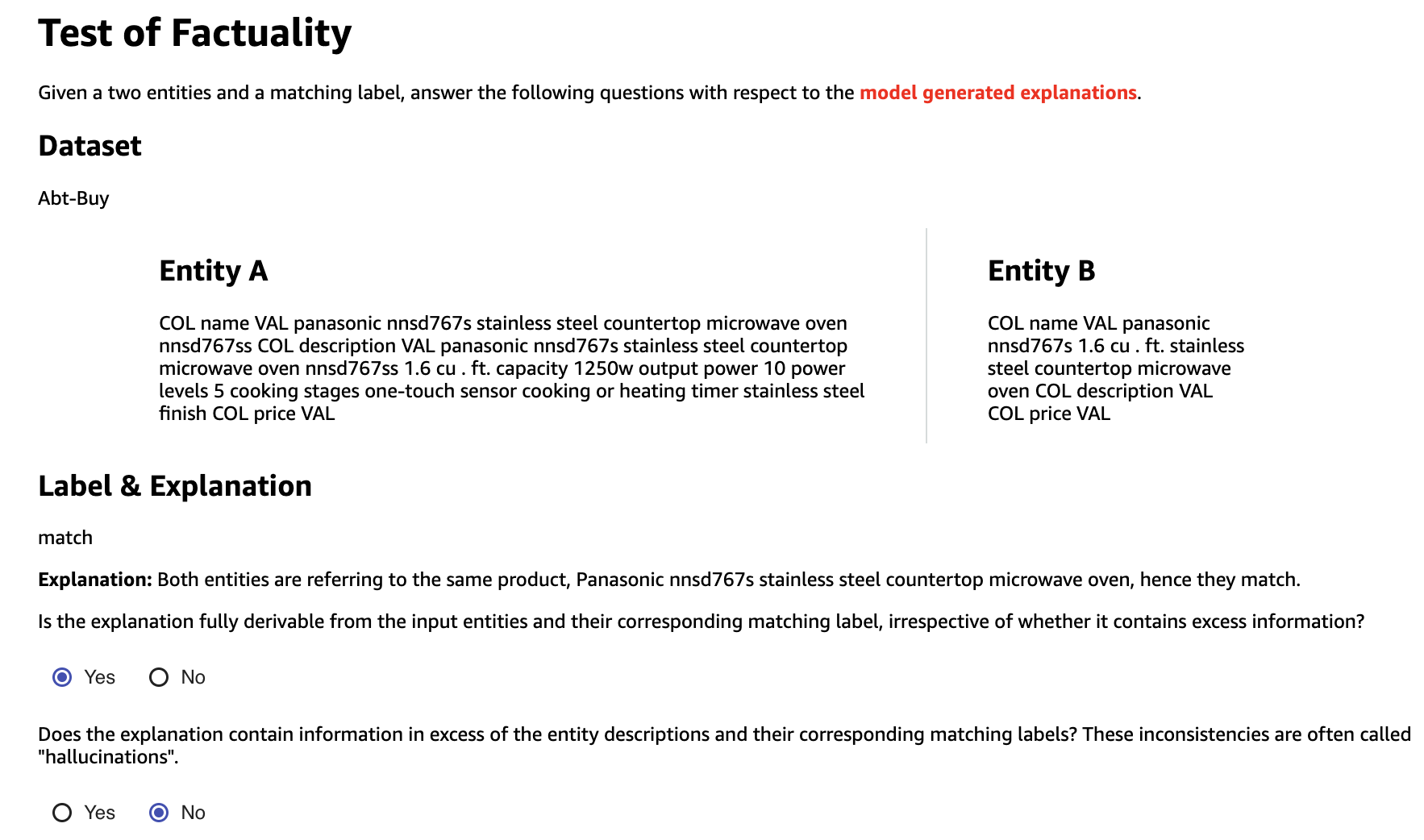}
  \caption{Interface to conduct Test of Factuality annotations on instances taken from the Abt-Buy dataset. Each model-generated (Mistral-7B; \citet{jiang2023mistral}) explanation is tested for intrinsic and extrinsic errors.}
  \label{appxfig:factuality_interface}
\end{figure*}

\section{Human Evaluation (H$_2$)}
\label{appx:human_eval}
We conduct Test of Factuality evaluation on Amazon Mechanical Turk (AMT) -- a popular platform for workers (both experts and non-experts) to perform ``micro-tasks'' (in our case, instance annotations) on explanations generated by the Mistral-7B model on 300 instances of the Abt-Buy dataset. Figure \ref{appxfig:factuality_interface} illustrates the interface provided to annotators where they're asked the two factuality-related questions and are presented with binary choices. 

\end{document}